\pdfoutput=1
\documentclass[11pt]{article}

\usepackage[final]{acl}

\usepackage{times}
\usepackage{latexsym}

\usepackage[T1]{fontenc}
\usepackage[utf8]{inputenc}

\usepackage{microtype}

\usepackage{inconsolata}

\usepackage{graphicx}
\usepackage{kotex}
\usepackage{booktabs}
\usepackage{multirow}
\usepackage{xspace}
\usepackage{tcolorbox}
\usepackage{amsmath}
\usepackage{color}
\usepackage{makecell}
\usepackage{xurl}
\usepackage{subcaption}
\usepackage{xcolor, soul}
\usepackage{amssymb}
\usepackage{bbm}
\usepackage{amsmath}
\usepackage{arydshln}
\usepackage{float}

\title{Team HUMANE at AVeriTeC 2025: HerO 2 for Efficient Fact Verification}

\author{
Yejun Yoon$^{\heartsuit}$~~~~~~Jaeyoon Jung$^{\clubsuit\diamondsuit}$~~~~~~Seunghyun Yoon$^{\spadesuit}$~~~~~~Kunwoo Park$^{\clubsuit\heartsuit}$\\ 
$^{\heartsuit}$Department of Intelligent Semiconductors, Soongsil University\\
$^{\clubsuit}$School of AI Convergence, Soongsil University\\
$^{\diamondsuit}$MAUM AI Inc.\\
$^{\spadesuit}$Adobe Research, USA\\
\texttt{\{yejun0382, jaeyoonskr\}@soongsil.ac.kr}, \texttt{syoon@adobe.com}, \texttt{kunwoo.park@ssu.ac.kr}
}

\begin{document}
\maketitle
\begin{abstract}
This paper presents HerO 2, Team HUMANE's system for the AVeriTeC shared task at the FEVER-25 workshop. HerO 2 is an enhanced version of HerO, the best-performing open-source model from the previous year’s challenge. It improves evidence quality through document summarization and answer reformulation, optimizes veracity prediction via post-training quantization under computational constraints, and enhances overall system performance by integrating updated language model (LM) backbones. HerO 2 ranked second on the leaderboard while achieving the shortest runtime among the top three systems, demonstrating both high efficiency and strong potential for real-world fact verification. The code is available at \url{https://github.com/ssu-humane/HerO2}.
\end{abstract}

\section{Introduction}

This paper describes Hero 2, the fact verification system developed by Team HUMANE for the AVeriTeC shared task. Hero 2 is an improved version of HerO~\cite{yoon-etal-2024-hero}, which achieved the state-of-the-art performance among the open-source models in the last year's AVeriTeC shared task. The 2025 edition emphasizes efficient, reproducible, and open-source approaches to automated fact-checking. Two key changes distinguish this year's task setting. First, computational and time constraints prohibit the use of large language models with more than ten billion of parameters (e.g., Llama3 70B Instruct). Second, the evidence evaluation metric has shifted from Hungarian METEOR (a token-based metric) to Ev2R recall (a model-based metric), which requires the generation of more flexible and semantically coherent evidence. 

In alignment with these goals, Hero 2 enhances retrieval performance through document-based retrieval and summarization, and reconstructs answer texts based on the question. We further improve the verification process using AWQ~\cite{lin2024awq}, enabling higher accuracy while maintaining efficiency under the hardware constraints specified by the task. As a result, Hero 2 achieved second place on the leaderboard while exhibiting the shortest runtime among the top-performing models, demonstrating its efficiency and suggesting its potential for real-world fact verification.

\begin{figure*}[t]
    \centering
    \includegraphics[width=.99\linewidth]{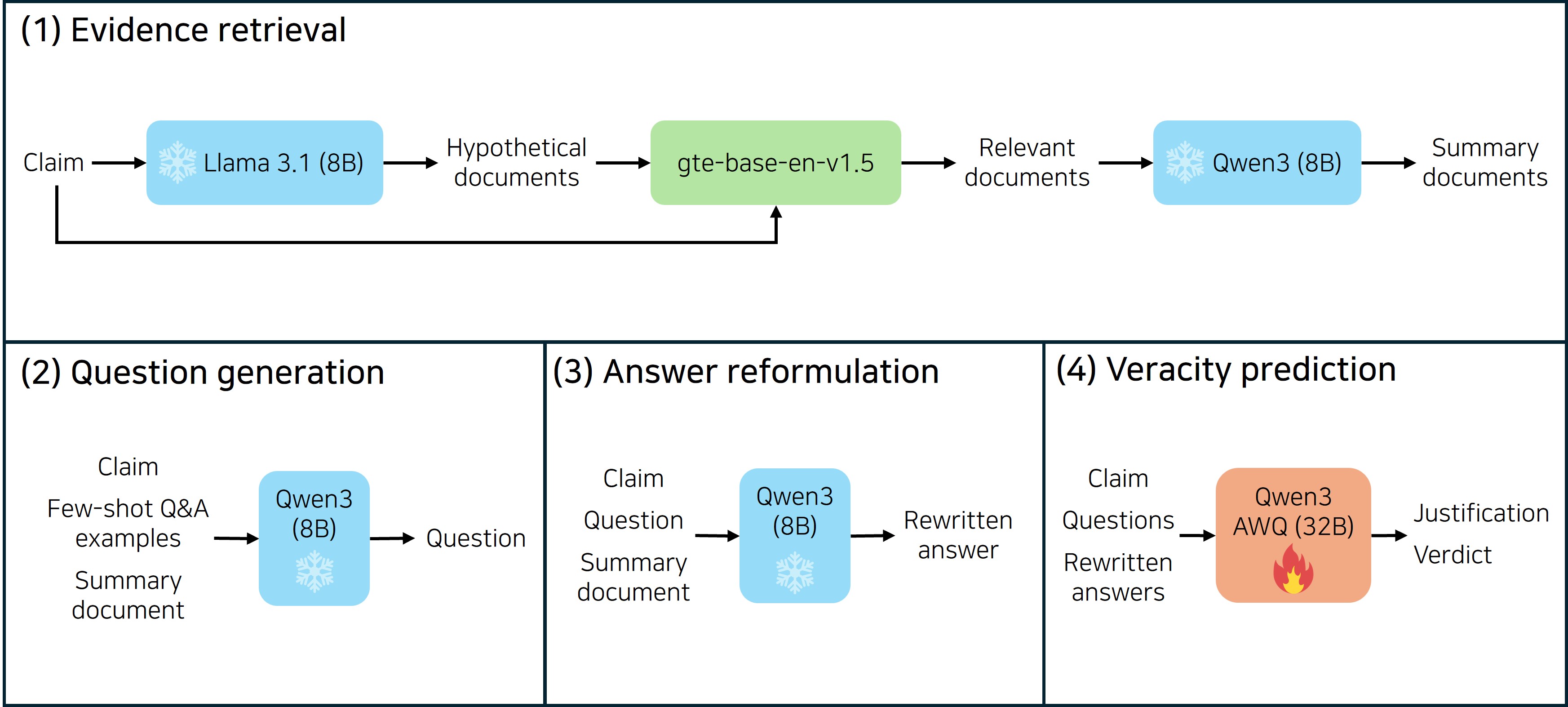}
    \caption{Pipeline of our system}
    \label{fig:pipeline}
\end{figure*}

\begin{table*}[t]
\small
\centering
\resizebox{.99\linewidth}{!}{
\begin{tabular}{cccccccc}
\toprule
System & \makecell{Query\\Expansion} & \makecell{Evidence\\Retrieval} & \makecell{Evidence\\Summarization} & \makecell{Question\\Generation} & \makecell{Answer\\Reformulation} & \makecell{Veracity\\Prediction}\\\midrule
 Baseline & \multirow{3}{*}{\makecell{HyDE-FC\\(Llama3.1 8B)}} & \makecell{Hybrid\\(BM25/SFR-embedding-2)} & NA & Llama3 8B & NA & Llama3.1 8B \\
 HerO 2 &  &  \makecell{Dense\\(gte-base-en-v1.5)} & Qwen3 8B & Qwen3 8B & Qwen3 8B & Qwen3 32B AWQ  \\\bottomrule
\end{tabular}}
\caption{Model configurations}
\label{tab:model_configuration}
\end{table*}

\section{Task Description}
The AVeriTeC shared task aims to build a fact-checking system that verifies real-world claims using web evidence. The claim verification process consists of three main steps. First, the system performs evidence retrieval by collecting relevant web documents. Next, during question generation, the system may generate questions for each piece of evidence to better assess the claim, though this step is optional. Finally, in the veracity prediction phase, the system uses the collected information to assess the truthfulness of the claim. The possible verdicts are: supported, refuted, not enough evidence, or conflicting evidence/cherry-picking. 

The 2025 shared task specifically targets two main goals. First, it aims to promote the development of high-performing systems that, using only open LLMs, can retrieve relevant evidence and generate accurate verdicts to maximize evaluation scores. Second, it emphasizes the importance of building reproducible and efficient fact-verification systems. Accordingly, all systems must be executable within the provided virtual machine environment and capable of verifying a single claim in under one minute. While the previous shared task~\cite{schlichtkrull-etal-2024-automated} used the Hungarian METEOR score to assess the quality of questions (Q  score) and question-answer pairs (Q+A score), this year’s evaluation adopts the Ev2R recall~\cite{akhtar2024ev2r}, an LLM-based evaluation method. Ev2R utilizes an LLM to decompose the ground-truth evidence into atomic facts. The Q+A score is then calculated by measuring the degree to which these facts cover the predicted evidence. The new AVeriTeC score is computed as the veracity prediction accuracy when the Q+A score of the predicted evidence for a claim exceeds a predefined threshold.

\section{Our System: HerO 2}

We present HerO 2, an improved fact verification pipeline of HerO~\cite{yoon-etal-2024-hero}. The key enhancements are summarized below:
\begin{itemize}
    \item \textbf{Document summarization}: Web documents are summarized into paragraph-level evidence blocks.
    \item \textbf{Answer reformulation}: A language model is prompted to convert the retrieved evidence blocks into answer-form texts.
    \item \textbf{Post-training quantization}: A fine-tuned LLM is quantized for veracity prediction.
    \item \textbf{Updated LM backbones}: Backbone LMs for each components are updated to maximize the performance.
\end{itemize} 

Figure~\ref{fig:pipeline} illustrates its overall pipeline, and Table~\ref{tab:model_configuration} details the model configuration in comparison to the baseline method.

\subsection{Knowledge Store Construction}
The 2025 AVeriTeC Shared Task imposes a one-minute time limit for processing each claim on a designated virtual machine. To meet this constraint, we apply two preprocessing steps to the web documents provided as the knowledge store: (1) indexing dense embeddings for all documents using gte-base-en-v1.5~\cite{li2023towards} following the design choice of the winning model in last year's shared task~\cite{rothermel-etal-2024-infact}; and (2) summarizing each document into paragraph-level evidence candidates using Qwen3 8B~\cite{yang2025qwen3}.

\subsection{Evidence Retrieval}
\label{evidence_retrieval}

\begin{figure}[t]
\begin{tcolorbox}[colback=white, fontupper=\small]
\textbf{Your task is to read the following document carefully and summarize it into a single, coherent paragraph. Focus on capturing the main ideas and essential details without adding new information or personal opinions.}\\

\textbf{Document:}: \textcolor{gray}{An image of a purported 1998 letter from actor Sean Connery (famous for his portrayal of agent James Bond) to Apple CEO Steve Jobs, caustically rebuffing an offer to become a pitchman for Apple Computers, hit the Internet in June 2011. ...}\\

\textbf{Output}: \textcolor{blue}{In June 2011, an image of a purported 1998 letter from actor Sean Connery to Apple CEO Steve Jobs, rejecting an offer to become an Apple pitchman, circulated online. However, the letter was part of a satirical article published on Scoopertino, a website known for fabricating news about Apple under the motto "All the News That's Fit to Fabricate." ...}

\end{tcolorbox}
\centering
\caption{An example of the instruction prompt used for document summarization, along with its output. The bold text is the instruction, and the blue text indicates the model output.}
\label{fig:document_summarization}
\end{figure}

The goal of evidence retrieval is to retrieve evidence necessary for verifying claims from the knowledge store. We use HyDE-FC~\cite{yoon-etal-2024-hero} for query expansion, which generates hypothetical fact-checking articles for a given claim by prompting an LLM. We retrieve the top 10 relevant documents through the indexed dense embedding. We adopt an additional step to summarize each of the retrieved documents into a single paragraph. The used prompt for summarization is shown in Figure~\ref{fig:document_summarization}.
Our best model uses Llama3.1 8B for HyDE-FC and Qwen3 8B for document summarization.

\subsection{Question Generation and Answer Reformulation}

\begin{figure}[ht]
\begin{tcolorbox}[colback=white, fontupper=\small]

\textbf{The following text provides an evidence obtained through web searches related to a specific question, used for verifying the accuracy of a claim.
Your task is to answer the question based on this evidence. Ensure your answer is Supported by relevant context from the evidence.}\\

\textbf{Claim}: \textcolor{gray}{In a letter to Steve Jobs, Sean Connery refused to appear in an apple commercial.}\\

\textbf{Question}: \textcolor{gray}{Was the letter from Sean Connery to Steve Jobs genuine?}\\

\textbf{Evidence}: \textcolor{gray}{In June 2011, an image of a purported 1998 letter from actor Sean Connery to Apple CEO Steve Jobs, rejecting an offer to become an Apple pitchman, circulated online. However, the letter was part of a satirical article published on Scoopertino, a website known for fabricating news about Apple under the motto "All the News That's Fit to Fabricate." ...}\\

\textbf{Answer}: \textcolor{blue}{No, the letter from Sean Connery to Steve Jobs was not genuine. It was part of a satirical article published on Scoopertino, a website known for fabricating news about Apple. ...}

\end{tcolorbox}
\centering
\caption{An example of the instruction prompt used for answer reformulation, along with its output. The bold text is the instruction, and the blue text indicates the model output.}
\label{fig:answer_generation}
\end{figure}

In this step, we generate questions based on the summarized evidence and retain only the information necessary to answer them through sequential LLM generations. We first generate questions using the same prompt as the baseline method. Then, we transform the retrieved summary text into an answer-form response conditioned on the claim and the generated question. The prompt used for this reformulation step is shown in Figure~\ref{fig:answer_generation}. Our best-performing pipeline employs Qwen3 8B for both question generation and evidence reformulation.

\subsection{Veracity Prediction}
We use a fine-tuned instruction-following language model to predict the veracity of given claims. Our best-performing model is a fine-tuned Qwen3 32B model, quantized to 4-bit using AWQ~\cite{lin2024awq} to satisfy the VRAM constraints of the shared task, enabling inference on an A10G GPU (23GB). Following the baseline approach, we use a prompt that incorporates the annotator's rationale into the veracity prediction process. The top 10 question–answer pairs generated in the earlier stages are provided as in-context examples, along with the claim to be verified.

\section{Evaluation Experiments}
This section presents the experimental results that guided the selection of each module in the submitted system.

\subsection{Experimental Setups}
In the comparison experiments, we used the development set to evaluate model performance. We used the training set for training our models. The Ev2R evaluation is carried out in a local environment using the Llama 3.3 70B~\cite{grattafiori2024llama} model with a threshold of 0.5.

For retrieval, we used gte-base-en-v1.5, which supports a context length of 8192. In other cases, we employed mxbai-embed-large-v1~\cite{emb2024mxbai}. All language models used in the experiments were instruction-tuned versions. For training, we used the Adam optimizer with a learning rate of 2e-5, batch size of 128, and trained the model for 2 epochs.

The Llama3.1 8B~\cite{grattafiori2024llama} was configured with the following hyperparameters for HyDE-FC: maximum number of tokens as 512, temperature as 0.7, and top-p as 1.0. The Qwen3 8B used the hyperparameters recommended by the Qwen team\footnote{\url{https://huggingface.co/Qwen/Qwen3-8B}}: temperature as 0.7, top-p as 0.8, top-k as 20, and min-p as 0. Veracity predictions used the hyperparameters: temperature as 0.9, top-p as 0.7, and top-k as 1. If the language model failed to produce a verdict label, we repeated the generation using top-2 sampling.

We ran experiments using three machines. The first has two H100 GPUs (80GB per GPU) and 480GB RAM. The second has eight H100 GPUs with 2TB RAM; the third has four NVIDIA A6000 GPUs (48GB per GPU) and 256GB RAM. The experiments were conducted in a computing environment with the following configuration: Python 3.12.9, PyTorch 2.6.0, Transformers 4.51.3, vLLM 0.8.5, and Sentence-Transformers 4.1.0.

\subsection{Experimental Results}

\paragraph{Evidence Retrieval}
We compare three retrieval strategies: (1) retrieving individual sentences, (2) retrieving consecutive sentence chunks with one-sentence overlap, and (3) retrieving entire documents. 
Table~\ref{tab:retrieve_exp} presents the evidence retrieval results on the development set, varying the number of retrieved evidence candidates. Among the three, the top-10 document-level retrieval strategy achieves the best performance, outperforming all other configurations with different retrieval targets and candidate counts.

\begin{table}[h]
\small
\centering
\begin{tabular}{lccc}
\toprule                
\multirow{2}{*}{Retrieval Target} & \multicolumn{3}{c}{Q + A (Ev2R recall)} \\
                        & Top-3  & Top-5 & Top-10 \\
\midrule
Sentence         & 0.289 & 0.315 & 0.374  \\
Chunk (2 sentences) & 0.311 & 0.369 & 0.404  \\
Chunk (3 sentences)    & 0.347 & 0.382 & 0.413  \\
Chunk (4 sentences)    & 0.364 & 0.41 & 0.115  \\
Document      & \textbf{0.37} & \textbf{0.487} & \textbf{0.522}  \\
\bottomrule
\end{tabular}
%}
\caption{Evidence retrieval performance}
\label{tab:retrieve_exp}
\end{table}

\paragraph{Answer Reformulation}
Table~\ref{tab:answer_reformulation} presents the results of document summarization and answer reformulation, varying the number of retrieved evidence documents used for veracity prediction. The best performance is observed when answer reformulation is applied to the top-10 question-answer pairs. When controlling for the number of retrieved documents, applying answer reformulation consistently outperforms the baseline without reformulation.

\begin{table}[h]
\small
\centering
\begin{tabular}{lccc}
\toprule
\multirow{2}{*}{Method} & \multicolumn{3}{c}{Q + A (Ev2R recall)}  \\
                        & Top-3 & Top-5 & Top-10 \\
\midrule
\makecell[l]{Document-based retrieval} & 0.37  & 0.487 & 0.522 \\
\makecell[l]{+ document summarization} & 0.483  & 0.51 & 0.487 \\
+ answer reformulation    & \textbf{0.501} & \textbf{0.514} & \textbf{0.556} \\
\bottomrule
\end{tabular}
\caption{Effects of answer reformulation}
\label{tab:answer_reformulation}
\end{table}

\paragraph{Veracity Prediction}
Controlling for other modules by fixing them to their best-performing configurations, we compare veracity prediction methods using the optimal settings for evidence retrieval and question generation. Table~\ref{tab:val_score} presents the fine-tuning results. While the models achieve comparable F1 scores, they exhibit substantial differences in accuracy. Since the AVeriTeC score is based on accuracy, we adopt it as the primary metric for selecting the best-performing model. Notably, applying AWQ post-training quantization to Qwen3 32B yields a +0.018 improvement in accuracy over its computationally equivalent smaller variant, Qwen3 8B. Among the 8B models, Qwen3 outperforms Llama 3.1, the language model used in the baseline.

\begin{table}[ht]
\small
\centering
\begin{tabular}{cccc}
\toprule
\multirow{1}{*}{Method} & \multirow{1}{*}{F1} & \multirow{1}{*}{ACC}\\
\midrule  
Qwen3 32B AWQ   & 0.382 & \textbf{0.692} \\
Qwen3 8B        & \textbf{0.385} & 0.674 \\
Llama3.1 8B     & 0.384 & 0.588 \\ \bottomrule
\end{tabular}
\caption{Veracity prediction performance}
\label{tab:val_score}
\end{table}

\subsection{Test Set Results} 

\begin{table}[ht]
\small
\centering
\resizebox{.99\linewidth}{!}{
\begin{tabular}{ccccc}
\toprule
System & AVeriTeC score & Average runtime per claim (s)\\\midrule
CTU AIC & \textbf{0.332±0.002} & 53.67\\
HerO 2 & 0.271±0.004 & \textbf{29.19} \\
yellow\_flash & 0.253±0.005 & 31.71 \\
Baseline  & 0.202±0.007 & 33.88 \\
\bottomrule
\end{tabular}
}
\caption{Test set results}
\label{tab:test_results}
\end{table}

Table~\ref{tab:test_results} presents the performance of HerO 2 on the test set in comparison with the baseline and other competitive models. CTU AIC achieved the highest AVeriTeC score of 0.332, followed by HerO 2 with 0.271 and yellow\_flash with 0.253. HerO 2 also achieved the lowest average runtime per claim at 29.19 seconds, demonstrating superior efficiency compared to CTU AIC (53.67 seconds), yellow\_flash (31.71 seconds), and the baseline (33.88 seconds). These results indicate that HerO 2 is the most efficient system among the top-performing models. It is worth noting that while some systems achieved shorter runtimes, their performance was significantly lower.

\section{Conclusion}
In this paper, we presented HerO 2, an efficient fact-checking system developed for the AVeriTeC shared task, hosted by the eighth FEVER workshop. Four key components contribute to the performance improvement of HerO 2: document summarization, answer reformulation, post-training quantization, and updated language model backbones. Our system achieved second place in the shared task while recording the shortest runtime among the top three systems, highlighting its efficiency and potential for real-world fact verification.

\section*{Acknowledgments}
This research was supported by the MSIT(Ministry of Science and ICT), Korea, under the Innovative Human Resource Development for Local Intellectualization (IITP-2025-RS-2022-00156360) and Graduate School of Metaverse Convergence support (IITP-2025-RS-2024-00430997) programs, supervised by the IITP(Institute for Information \& Communications Technology Planning \& Evaluation). Yejun Yoon and Jaeyoon Jung contributed to this work equally as co-first authors. Kunwoo Park is the corresponding author.

\bibliography{custom}

\end{document}